\documentclass{article}

\PassOptionsToPackage{numbers, compress}{natbib}

\usepackage[preprint]{neurips_2020}

\usepackage{color}
\usepackage{verbatim}
\usepackage[utf8]{inputenc} 
\usepackage[T1]{fontenc}    
\usepackage{hyperref}       
\usepackage{url}            
\usepackage{booktabs}       
\usepackage{amsfonts}       
\usepackage{amsmath}
\usepackage{nicefrac}       
\usepackage{microtype}      
\usepackage{graphicx}
\usepackage[tight]{subfigure}
\newtheorem{theorem}{Theorem}
\usepackage{bm}

\title{ProbAct: A Probabilistic Activation Function \\ for Deep Neural Networks}

\author{Kumar~Shridhar \ $^\spadesuit$, Joonho~Lee \ $^\clubsuit$ , Hideaki~Hayashi \ $^\clubsuit$, \textbf{Purvanshi Mehta} \ $^\bullet$\\ , \textbf{Brian Kenji Iwana }\ $^\clubsuit$ , \textbf{Seokjun Kang} \ $^\clubsuit$ , \textbf{Seiichi Uchida} \ $^\clubsuit$ , \textbf{Sheraz Ahmed} \ $^\diamond$ \\ , \textbf{Andreas Dengel} \ $^\diamond$\\
\\
$^\spadesuit$ University of Kaiserslautern, Germany
$^\clubsuit$ Kyushu University, Japan\\
$^\bullet$ University of Rochester, USA
$^\diamond$ DFKI, Kaiserslautern \\
}

\begin{document}
\maketitle

\footnote{
\\$^\spadesuit$ \texttt{k\_shridhar16@cs.uni-kl.de} \\
$^\clubsuit$ \texttt{\{joonho.lee, hayashi, brian, seokjun.kang,  uchida\} @human.ait.kyushu-u.ac.jp}  \\
$^\bullet$ \texttt{pmehta9@ur.rochester.edu}\ \
$^\diamond$ \texttt{\{sheraz.ahmed, andreas.dengel\}@dfki.de}
}
\begin{abstract}

Activation functions play an important role in training artificial neural networks. The majority of currently used activation functions are deterministic in nature, with their fixed input-output relationship. In this work, we propose a novel probabilistic activation function, called \textit{ProbAct}. ProbAct is decomposed into a mean and variance and the output value is sampled from the formed distribution, making ProbAct a stochastic activation function. The values of mean and variances can be fixed using known functions or trained for each element. In the trainable ProbAct, the mean and the variance of the activation distribution is trained within the back-propagation framework alongside other parameters. We show that the stochastic perturbation induced through ProbAct acts as a viable generalization technique for feature augmentation. In our experiments, we compare ProbAct with well-known activation functions on classification tasks on different modalities: Images (CIFAR-10, CIFAR-100 and STL-10) and Text (Large Movie Review). We show that ProbAct increases the classification accuracy by +2-3\% compared to ReLU or other conventional activation functions on both original datasets and when datasets are reduced to 50\% and 25\% of the original size. Finally, we show that ProbAct learns an ensemble of models by itself that can be used to estimate the uncertainties associated with the prediction and provides robustness to noisy inputs.

\end{abstract}

\section{Introduction}
Activation functions add non-linearity to neural networks making them learn complex functional mappings from data~\cite{article2}. Different activation functions with different characteristics have been proposed. Sigmoid~\cite{Cybenko1989} and hyperbolic tangent (Tanh) were the popular ones during the early usage of neural networks~\cite{schmidhuber2015deep} mainly due to their monotonicity, continuity, and bounded properties. In recent times, the Rectified Linear Unit (ReLU)~\cite{nair2010rectified} has become an extremely popular activation function for neural networks. Several variants of ReLU have been proposed, e.g., Leaky ReLU~\cite{xu2015empirical}, Parametric ReLU (PReLU)~\cite{he2015delving}, and Exponential Linear Unit (ELU)~\cite{2015arXiv151107289C}.

However, all of these are deterministic activation functions with fixed input-output relationships.
In this work, we propose a new activation function, called \textit{ProbAct}, which is not only trainable but also stochastic in nature. The idea of ProbAct is inspired by the stochastic behavior of biological neurons. 
Noise in neuronal spikes can arise due to uncertain bio-mechanical effects~\cite{lewicki1998review}. We try to emulate a similar behavior in the information flow to the neurons by injecting stochastic sampling from a Gaussian distribution to the activations. Consequently, even for the same input value $x$, the output value from ProbAct varies stochastically --- a capability conventional activation functions do not offer. 

The induced perturbations prove to be effective in avoiding overfitting during training, thus yielding better generalizations. Since the operation is a resemblance to feature augmentation, we call it \emph{augmentation-by-activation}. Furthermore, we show that ProbAct improves the classification accuracy by 2-3\% compared to ReLU or other conventional activation functions on established image datasets and 1-2\% on text datasets. The main contributions of our work are as follows: 
\begin{itemize}
    \item We introduce a novel activation function, called ProbAct, whose output undergoes stochastic perturbation. 
    \item We propose a novel method of governing the stochastic perturbation with parameters trained through back-propagation.
    \item We show that ProbAct improves the performance on various visual and textual classification tasks while generalizing well on reduced datasets.
    \item We also show that the improvement by ProbAct is realized by the augmentation-by-activation, which acts as a stochastic regularizer to prevent overfitting of the network and acts as a feature augmentation method.
    \item Finally, we demonstrate that ProbAct learns an ensemble of models by itself, allowing the estimation of predictive uncertainties and robustness to noisy data.

\end{itemize}
 
\begin{figure}
\centering
\subfigure[ReLU]{\includegraphics[height=.25\linewidth]{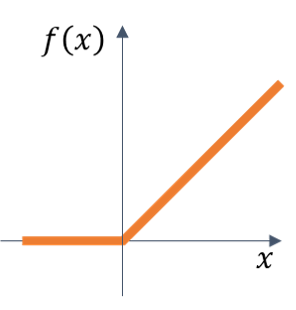}} \hfill 
\subfigure[ProbAct]{\includegraphics[height=.25\linewidth]{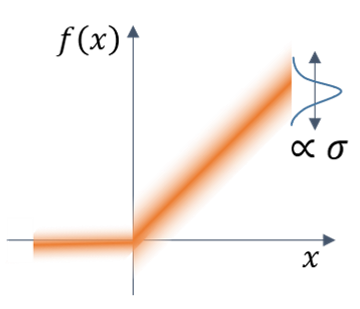}} \hfill
\subfigure[Effect of stochastic perturbation]{\includegraphics[height=.23\linewidth]{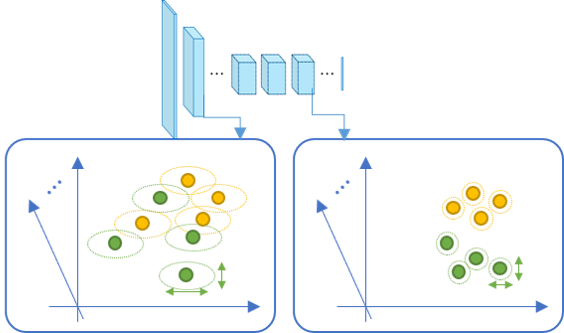}}
\caption{Comparison of (a) ReLU and (b) the proposed activation function. (c) is the effect of stochastic perturbation by ProbAct at feature spaces in a neural network.}
\label{fig:probact}
\end{figure}

\section{Related Work}
\subsection{Activation Functions}

Various approaches have been applied in the past to create a desired activation function. 
Research on activation functions can be broadly separated into two approaches: fixed activation functions, and adaptive activation functions. 

Fixed activation functions are constant pre-determined functions such as sigmoid, tanh, and ReLU. 
In particular, ReLU has led to significant improvements in neural network performance. However, ReLU faces a dead neuron problem and variants of ReLU were suggested to solve this problem. 
For example, Leaky ReLUs~\cite{xu2015empirical} use a fixed $0.01x$ value for $x<0$. 
Other activation functions like Swish~\cite{Ramachandran2018SearchingFA} and Exponential Linear Sigmoid SquasHing (ELiSH)~\cite{Basirat2018TheQF} take a different approach and use bounded negative regions. 

Adaptive activation functions use trainable parameters in order to optimize the activation function. 
For example, Parametric ReLUs (PReLUs)~\cite{he2015delving} are similar to Leaky ReLUs but with a trainable parameter instead of a fixed value. 
In addition, S-shape ReLU (SReLU)~\cite{Jin:2016:DLS:3016100.3016142} and Parametric ELU (PELU)~\cite{trottier2017parametric} were suggested to improve the performance of conventional ReLU functions. 

\subsection{Generalization and Stochastic Methods}

There has been less research on stochastic activation functions due to expensive sampling processes. Noisy activation functions~\cite{pmlr-v48-gulcehre16} tried to deal with these problems by adding noises to the non-linearity in proportion to the magnitude of saturation of the non-linearity. 
RReLU~\cite{xu2015empirical} uses Leaky ReLUs with randomized slopes during training and a fixed slope during testing.

There are many other generalization techniques that use random distributions or stochastic functions. 
For example, dropout~\cite{srivastava2014dropout} generalizes by removing connections at random during training and can be interpreted as a way of model averaging. 
\cite{liu2018towards} suggested Random Self-Ensemble (RSE) by combining randomness and ensemble learning, and 
\cite{Inayoshi} proposed back-propagating noise to the hidden layers. 
Furthermore, the effects of adding noise to inputs~\cite{Bishop_1995,An_1996}, the gradient~\cite{Audhkhasi_2013,neelakantan2015adding}, and weights~\cite{An_1996,NIPS2011_4329,Murray_1993,pmlr-v37-blundell15} have been studied.
However, we adopt the concept that stochastic neurons with sparse representations allow internal regularization as shown by ~\cite{bengio2013estimating}.

Neural networks using point estimates as weights and fixed activations always result in the same prediction with no uncertainty estimates.
Introducing Bayesian inference on weights of a network \cite{mackay1992practical,graves2011practical,hinton1993keeping,shridhar2019comprehensive} is one way to estimate the predictive uncertainty. \cite{kendall2017uncertainties,kwon2018uncertainty,shridhar2018uncertainty} show other approaches to estimate the predictive uncertainty and decompose it according to its origin. However, Bayesian neural networks increase the number of parameters as weights are represented by means of a parametric model. Deep ensembles \cite{lakshminarayanan2017simple} uses non-Bayesian networks to achieve similar results but training $N$ networks for any large task is computationally very expensive. We show that ProbAct acts as ensembles of networks that can be used to estimate predictive uncertainties with only a few additional parameters. 

\cite{DBLP:journals/corr/WangSY16} proposed natural parameter networks (NPN), where exponential-family distributions represented the inputs, targets and weights. \cite{DBLP:journals/corr/abs-1805-11327} and  \cite{hernandez2015probabilistic} also treat the output of activation function as distributions rather than just deterministic values. However, ProbAct's variance is input-independent while \cite{DBLP:journals/corr/WangSY16, DBLP:journals/corr/abs-1805-11327,hernandez2015probabilistic} produce input-dependent variance, making ProbAct faster and less computationally expensive. 

\section{ProbAct: A Stochastic Activation Function}
Every layer of a neural network computes its output $y$ for the given input $\bm{x}$: 
\begin{equation}
y = f(\bm{w}^\mathrm{T} \bm{x}), 
\end{equation}
where $\bm{w}$ is the weight vector of the layer and $f(\cdot)$ can be any activation function, such as ProbAct.
ProbAct is defined as:
\begin{equation}
f(\bm{x}) = \mu(\bm{x}) + z,
\label{eq:probact}
\end{equation}
where $\mu(\bm{x})$ is a static or learnable mean (for example, $\mu(\bm{x})=\max(0,\bm{x})$ if it is static ReLU) and the perturbation term $z$ is: 
\begin{equation}
z = \sigma\epsilon.
\label{eq:mu def}
\end{equation}
The perturbation parameter $\sigma$ is a fixed or trainable value which specifies the range of stochastic perturbation and $\epsilon$ is a random value sampled from a normal distribution $\mathcal{N}(0,1)$.
The value of $\sigma$ is either determined manually or trained along with other network parameters (i.e., weights) with simple implementation. With decreasing $\sigma$, ProbAct converges to its mean function $\mu(\bm{x})$. If $\sigma\to 0$, ProbAct behaves the same as its mean function. Example: if the mean is a fixed ReLU function, then ProbAct acts a generalization of ReLU in that case. 

\subsection{Setting the Parameter for Mean}

The mean function $\mu(\bm{x})$ is trained for every input $\bm{x}$, i.e. element-wise. However, learning the mean value with zero or random initialization takes unnecessarily long to converge. So, we propose ways for mean initialization. 

\subsubsection{Mean initialization:}

We propose initialization of $\mu$ with known functions such as ReLU with $\mu(\bm{x})=\max(0,\bm{x})$. Besides ReLU, any known functions can be used as an initializer. We use ReLU for its simplicity and good convergence behavior. 

\subsection{Setting the Parameter for Stochastic Perturbation}
The parameter $\sigma$ specifies the range of stochastic perturbation. In the following, we will consider two cases of setting $\sigma$, fixed and trainable. 

\subsubsection{Fixed Case}
There are several ways to choose the desired $\sigma$. The simplest is setting $\sigma$ to be a constant hyper-parameter. Choosing one constant $\sigma$ for all elements  is theoretically justified as $\epsilon$ is randomly sampled from a normal distribution and $\sigma$ acts as a scaling factor to the sampled value $\epsilon$. This can be interpreted as repeated addition of the scaled Gaussian noise to the activation maps, which helps in better convergence of the network parameters \cite{bengio2013estimating}. The network is optimized using gradient-based learning.

The nature of $z = \sigma \epsilon$ is Gaussian as $\epsilon\sim\mathcal{N}(0,1)$, and $\sigma$ being a constant value which does not affect the Gaussian properties. This ensures learning using gradient-based methods. The proposed method does not significantly affect the number of parameters in the architecture, hence comes at no additional computational cost. However, choosing the best $\sigma$ is a difficult task as setting any other hyper-parameter for training a neural network.

\subsubsection{Trainable Case}

Using a trainable $\sigma$ reduces the requirement to determine $\sigma$ as a hyper-parameter and allows the network to learn the appropriate range of sampling.
There are two ways of introducing a trainable $\sigma$:
\begin{itemize}
    \item \textit{Single Trainable $\sigma$}: A shared trainable $\sigma$ across the network. This introduces a single extra parameter used for all ProbAct layers. This is similar to the fixed $\sigma$ but the value is trained. 
    \item \textit{Element-wise Trainable $\sigma$}: This method uses a trainable parameter for each input element. This adds the flexibility to learn a different distribution for every input-output mapping. 
\end{itemize}

\paragraph{Training $\sigma$}

The trainable parameter $\sigma$ is trained using a back-propagation simultaneously with other model parameters. The gradient computation of $\sigma$ is done using the chain rule. 
Given an objective function $\mathcal{E}$, the gradient of $\mathcal{E}$ with respect to $\sigma_{l,i}$, where $\sigma_{l,i}$ is the perturbation parameter and $y_{l,i}$ is the output of the $i$-th unit in the $l$-th layer, is:
\begin{equation}\label{eq:gradSigma}
\frac{\partial \mathcal{E}}{\partial \sigma_{l,i}} = \frac{\partial \mathcal{E}}{\partial y_{l,i}}\frac{\partial y_{l,i}}{\partial \sigma_{l,i}},
\end{equation}
The term $\frac{\partial \mathcal{E}}{\partial y_{l,i}}$ is the gradient propagated from the deeper layer.
The gradient of the activation is given by:
\begin{equation}
\frac{\partial y_{l,i}}{\partial \sigma_{l,i}} = \epsilon.
\end{equation}

\paragraph{Bounded $\sigma$}

Training $\sigma$ without any bounds can create perturbations in a highly unpredictable manner when $\sigma \rightarrow \infty$, making training difficult. Taking the advantages of monotonic nature of the sigmoid function, we bound the upper and lower limit of $\sigma$ to $(0,\alpha)$ using:
\begin{equation}
\sigma = \alpha\;\mathrm{sigmoid}(\beta k),
\label{eq:sigma}
\end{equation}
where $k$ represents the element-wise learnable parameter, and $\alpha$ and $\beta$ are scaling parameters that can be set as hyper-parameters.
 
\subsection{Stochastic Regularizer}

Figure \ref{fig:probact} (c) illustrates the effect of stochastic perturbation by ProbAct in the feature space of each neural network layer. 
Intuitively, ProbAct adds perturbation to each feature vector independently, 
and this function acts as a regularizer to the network. 
It should be noted that while the noise added to each ProbAct is isotropic, the noise from early layers is propagated to the subsequent layers; hence, the total noise added to a certain layer depends on the noise and weights of the early layers.

Further, the effect of regularization is proportional to the variance of the distribution. A high variance is induced with a higher $\sigma$ value, allowing sampling from a high variance distribution which is further away from the mean. This way the prediction is not over-reliant on one value, helpful in countering overfitting problems. For the fixed $\sigma$ case, the variance of the noise is constant. However, it helps in optimizing the weights of the network.

\section{Experiments}
In the experiments, we empirically evaluate ProbAct on image classification and sentiment analysis tasks to show the effectiveness of the proposed method. 

\paragraph{Datasets}

To evaluate our proposed activation function, we use three image classification datasets, CIFAR-10~\cite{cifar10}, CIFAR-100~\cite{cifar10}, and STL-10~\cite{coates2011analysis} and one text dataset: Large Movie Review \cite{maas-EtAl:2011:ACL-HLT2011}. More information on the dataset distribution is mentioned in the Appendix. 

\begin{table} 
  \caption{Performance comparison of various activation functions and a Variation Inference Neural Network with ProbAct. The test accuracy(\%) indicates the average of testing over three runs. The train and test time is reported with respect to ReLU activation function and is measured in seconds and milliseconds respectively.}
  \label{acc-table}
  \centering
  \begin{tabular}{lcccccc}
    \toprule                 
    Activation function & CIFAR-10 & CIFAR-100 & STL-10 & IMDB & Train time & Test time\\
     &  &  &  &  & (sec) & (milli-sec)\\
    \midrule
    Sigmoid & 10.00  & 1.00 & 10.00 & 85.92 & 1.07 & 1.03\\
    Tanh & 10.00 & 1.00 & 10.00 & 85.88 & 1.08 & 1.03\\
    ReLU & 87.27  & 52.94 & 60.80 & 85.85 &\textbf{1.00} & \textbf{1.00}\\
    Leaky ReLU & 86.49 & 49.44 & 59.16 &  85.47 &1.04 & 1.08 \\
    PReLU & 86.35 & 46.30 & 60.01 & 85.95 &1.16 & \textbf{1.00} \\
    ELU & 87.65 & 56.60 & 64.11 & 86.51 &1.16 & 1.04 \\
    SELU & 86.65 & 51.52 & 60.71 & 85.71 & 1.19 & 1.05 \\
    Swish & 86.55 & 54.01 & 63.50 & 86.14 &1.20 & 1.13   \\
    \midrule
    Bayesian VGG VI\footnotemark & 86.22 & 48.27 & 57.22 & - & - & -   \\
    \midrule
    {ProbAct} \\
    \quad\quad \textbf{Mean}  &  & & & &  & \\
    \quad Element-wise $\mu$  & 85.80 & 48.50 &  54.17 & 83.86 & 1.29 & 1.35\\
    \quad\quad \textbf{Sigma}  &  & & & &  & \\
    \quad Fixed ($\sigma = 0.5$ ) & 88.50  & 56.85 & 62.30 & \textbf{87.31} & 1.09 & 1.25\\
    \quad Fixed ($\sigma = 1.0$)& 88.87  & \textbf{58.45} & 62.50 & 87.00 & 1.10 & 1.27 \\
    \quad One Trainable $\sigma$ & 87.40 & 53.87 & 63.07 & 86.35 & 1.23 & 1.30\\ 
    \quad EW Trainable $\sigma$ &  & & & &  & \\
    \quad \quad Unbound & 86.40 & 54.10 & 61.70 & 86.64 & 1.25 & 1.31 \\
    \quad  \quad Bound &\textbf{88.92} & 55.83&  \textbf{64.17} & 85.86 & 1.26 & 1.33 \\
    \bottomrule
  \end{tabular}
\end{table}
\footnotetext{Results taken from this implementation :\\  \url{https://github.com/kumar-shridhar/PyTorch-BayesianCNN}}

\subsection{Experimental Setup}
To evaluate the performance of the proposed method on classification tasks, we compare ProbAct to the following activation functions: ReLU, Sigmoid, Hyperbolic Tangent (TanH), Leaky ReLU~\cite{xu2015empirical}, PReLU~\cite{he2015delving}, ELU~\cite{clevert2015fast}, SELU~\cite{klambauer1706self}, Swish~\cite{Ramachandran2018SearchingFA} and to a Bayesian VGG network using variational inference \cite{Shridhar2018BayesianCN}. We utilize a 16-layer VGG neural network~\cite{2014arXiv1409.1556S} architecture for the image classification task and a two-layer CNN network for sentiment analysis task. 
The architecture, specific hyper-parameters, and training settings are provided in the Appendix section.

For a fair and consistent evaluation environment, we did not use regularization tricks, pre-training, and data augmentation to show the true comparison of the activations. The inputs are normalized to $[0,1]$. 
The STL-10 images are resized to 32 by 32 to match the CIFAR datasets to keep a fixed input shape to the network. 

For trainable mean, we keep $\sigma = 1$ to see the effects of the learned $\mu$. For our experiments, we train $\mu$ element-wise with an initialization of $\mu(\bm{x})=\max(0,\bm{x})$. Initializing mean with $\max(0,\bm{x})$ showed faster convergence when compared with zero or random mean initialization.

For  evaluations of $\sigma$, we keep $\mu(\bm{x})=\max(0,\bm{x})$ fixed and we report fixed $\sigma$ values of $0.5$ and $1.0$. 
In case of trainable $\sigma$, we set three types of $\sigma$ values: One Trainable $\sigma$, Element-wise (EW) Trainable $\sigma$ (unbound), and Element-wise Trainable $\sigma$ (bound). 
Element-wise Trainable $\sigma$ (bound) is the Element-wise Trainable $\sigma$ when $\sigma$ is bound by $\sigma = \alpha\;\mathrm{sigmoid}(\beta k)$ and Element-wise Trainable $\sigma$ (unbound) lacks this constraint. We used $\alpha = 2$ and $\beta = 5$ in the experiments. These values were found through exploratory testing.

We did not see any improvements while training both $\mu \  \text{and}\ \sigma$ together as their individual accuracy is similar and merging the two modalities did not help. However, training them individually have their own advantages as mentioned in the next section.

\begin{figure}
\centering
\includegraphics[width=1\linewidth]{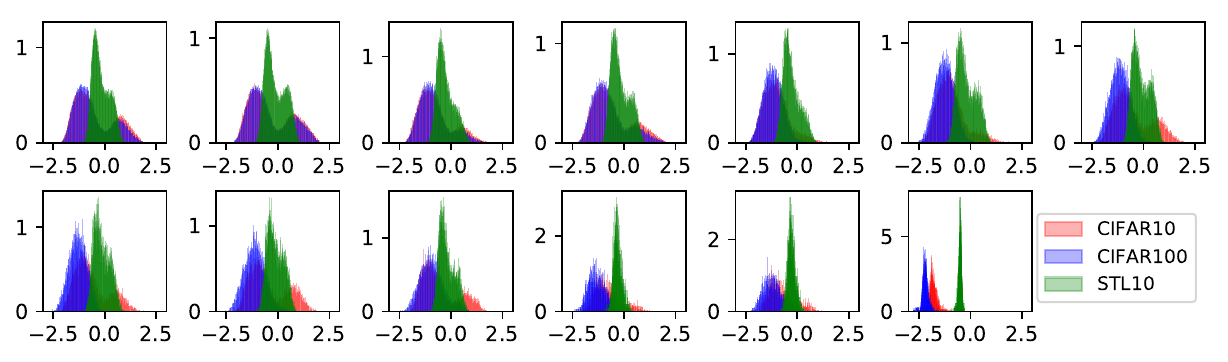}
\caption{Histograms show how element-wise learnable parameters, $k$ are distributed after training VGG-16 architecture for CIFAR-10, CIFAR-100 and STL-10 datasets at each layer. X-axis denotes $k$ value after training and Y-axis denotes its frequency. Every subfigure represents a ProbAct layer in layer-wise order from top left to bottom right.}
\label{fig:ProbActCIFARtest}
\end{figure}

\subsection{Quantitative Evaluation}
The results of the experiment on CIFAR-10, CIFAR-100, STL-10, and IMDB are shown in Table~\ref{acc-table}. 
These experiments are performed three times and the average of the three is reported in the results.  
When using Element-wise Trainable $\sigma$ (bound) ProbAct, we achieved performance improvements of 2.25\% on CIFAR-10, 2.89\% on CIFAR-100, 3.37\% on STL-10, and 1.5\% on IMDB datasets compared to the standard ReLU. 
In addition, the proposed method performed better than any of the evaluated activation functions. 

In order to demonstrate the applicability of our proposed method, the training and testing times relative to the standard ReLU are also shown in Table~\ref{acc-table}. 
The time comparison shows that we can achieve higher performance with only a relatively small time difference. 
This is mainly because the learnable $\sigma$ values are few compared to the learnable weight values in a network.
 Hence, our approach comes at nearly no additional time cost. This shows ProbAct as a strong replacement over popular activation functions. 

We visualized training aspects of the Single Trainable $\sigma$ in Figure~\ref{fig:comparisonNoise} (a) for 200 epochs for CIFAR-10 dataset. We trained the network for 400 epochs and cropped it for 200 epochs for better visualization as there is no significant change in the $\sigma$ value after 200 epochs. After 100 training epochs, the Single Trainable $\sigma$ goes towards 0. 
However, in order to test the Single Trainable $\sigma$ when $\sigma=0$, we replaced ProbAct with ReLU on the trained network. 
We confirmed that even with ReLU on the network trained with Single Trainable $\sigma$, we could achieve higher results than when training on ReLU. 
This shows that while training, $\sigma$ helps to optimize the other learnable weights better than standard ReLU architecture, allowing better model performance.
Figure~\ref{fig:comparisonNoise} (b) shows the mean Element-wise Trainable $\sigma$ over 200 epochs for all the layers. We demonstrate the ability of the network to train element-wise $\sigma$ across all layers, even when the number of trainable parameters is increased due to Element-wise Trainable $\sigma$ parameters.

Figure~\ref{fig:ProbActCIFARtest} shows the frequency distribution for the bounded element-wise trained $k$ values after 400 epochs. We observed two peak values for every distribution across all three image datasets. We assume that the derivative of a sigmoid function becomes 0 at both the boundaries of the function. The points in the left peak lie in the lower boundary of $\sigma$ (0 in our case) making ProbAct behave as ReLU. Right peak points lie in the upper boundary of the sigmoid (2 in our case) and take $\sigma$ as 2. The values in between the peaks signify other $\sigma$ values. 

In the case of both CIFAR datasets, the distribution of parameter $k$ in the last layer is quite narrow and concentrated in the negative domain. As shown in Figure~\ref{fig:comparisonNoise}~(b), the $\sigma$ values becomes 0, which indicates that ProbAct conducts the ReLU-like operation. 


\subsection{Augmentation by Activation}
\begin{figure}
\subfigure[One Trainable $\sigma$]{\includegraphics[width=0.24\linewidth]{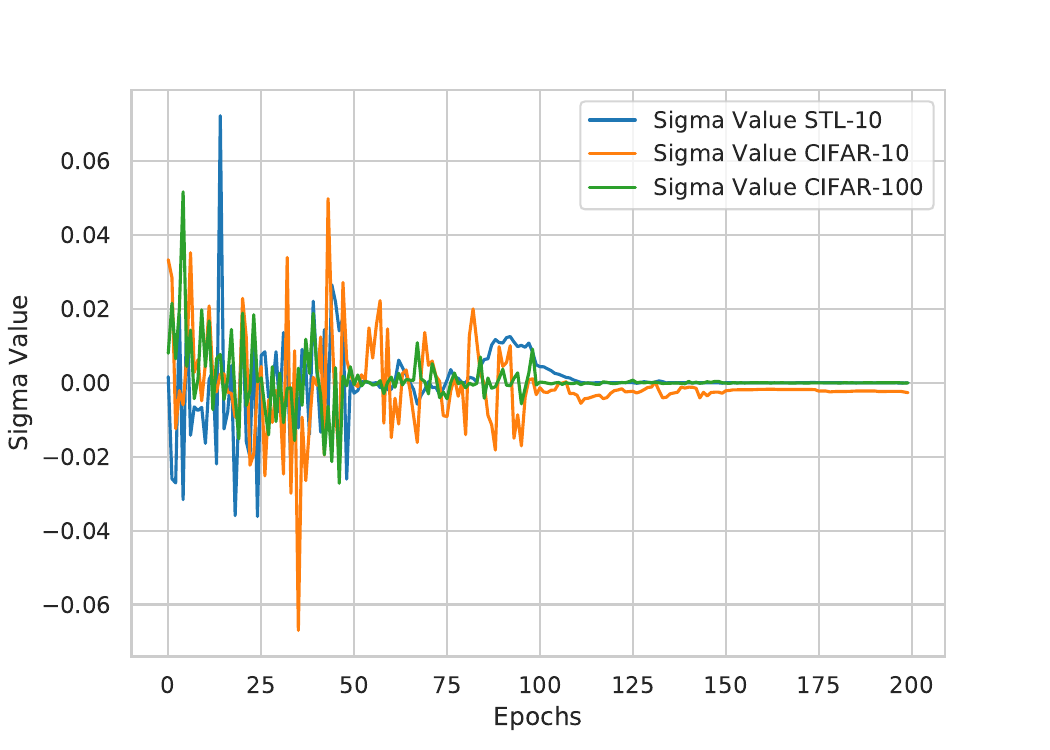}}\hfill
\subfigure[Layer Wise $\sigma$ Value]{\includegraphics[width=0.25\linewidth]{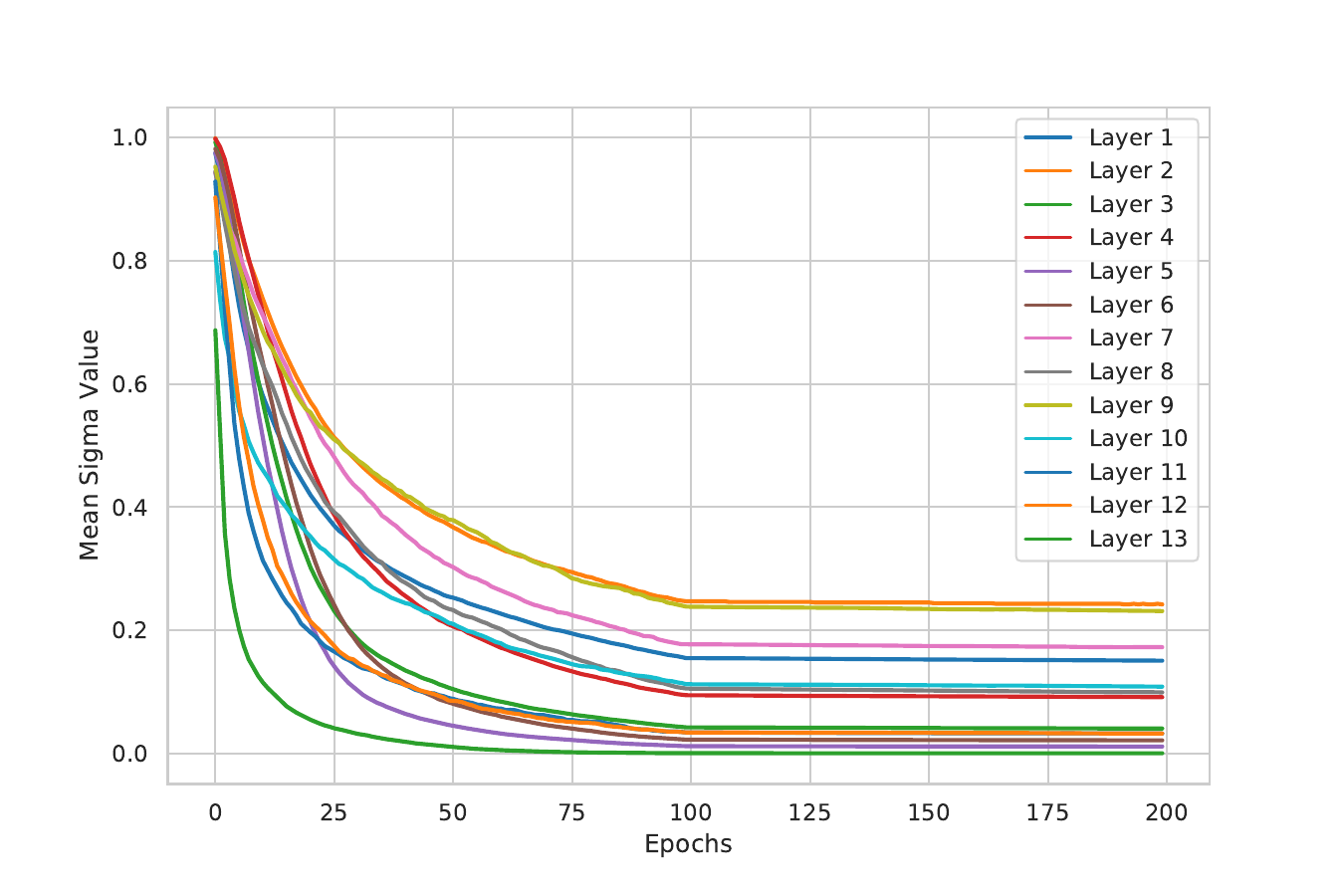}}\hfill
\subfigure[Noisy Input]{\includegraphics[width=0.25\linewidth]{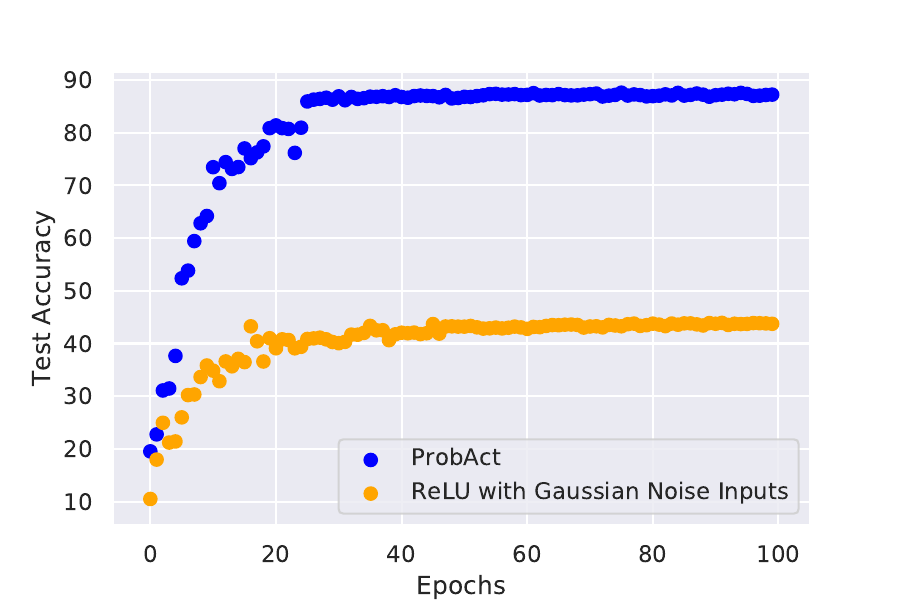}}\hfill
\subfigure[Noisy Activation]{\includegraphics[width=0.25\linewidth]{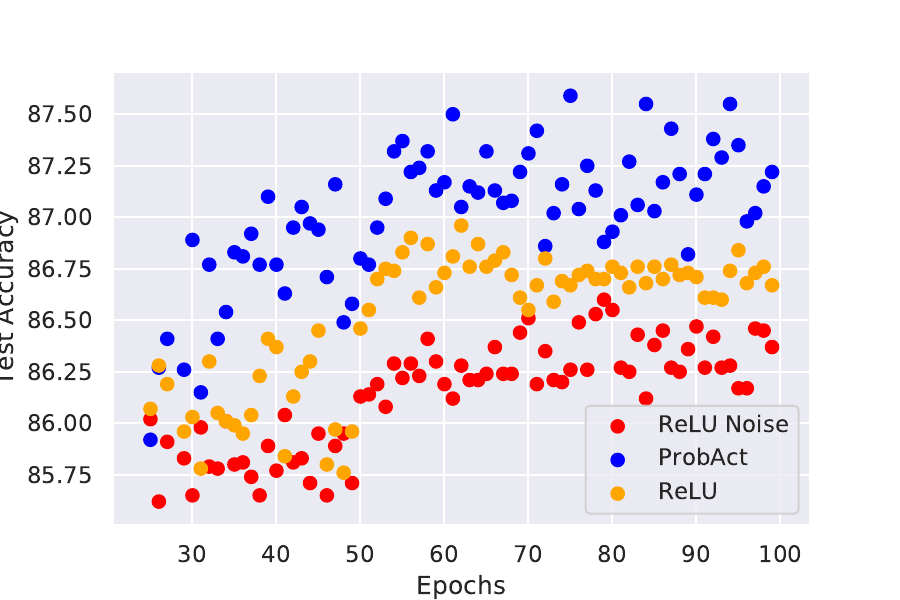}}
\caption{(a) shows the transition of single trainable $\sigma$ for VGG-16 architecture on the three datasets. (b) shows the Layerwise mean $\sigma$ value for VGG-16 ProbAct layers trained on CIFAR10 dataset. (c) shows the test accuracy comparison between ReLU with noisy input and ProbAct, while (d) shows the test accuracy comparison of ReLU, ReLU with noisy activation and ProbAct on CIFAR-10 dataset.}
\label{fig:comparisonNoise}
\end{figure}

We add ProbAct to the first layer of the VGG network keeping ReLU as the activation function for all the other layers. We show that perturbation induced by ProbAct in the first layer behaves as augmentation added in the activation function; improving the overall network generalization ability. ProbAct can be thought of as a way to add an adaptable and trainable perturbation enhancing the generalization capability of the network. These added perturbations are different than just adding noise to either the activation or to the inputs. Figure~\ref{fig:comparisonNoise} (c) draws a comparison between ProbAct in the first layer with noisy input (in our experiments, Gaussian noise was added to the inputs sampled from a distribution $\mathcal{N}(0,1)$), while in Figure~\ref{fig:comparisonNoise} (d), Gaussian noise is added to the first activation layer. ProbAct with learnable variance outperforms both the standard ReLU with noisy inputs and standard ReLU with noisy activations. Our proposed method adopts stochastic noise into the activation function in a controlled manner. To the best of our knowledge, our proposed method is the first approach that adopts learnable stochastic noise into activation function.

\begin{figure}
\subfigure[]{\includegraphics[width=0.25\linewidth]{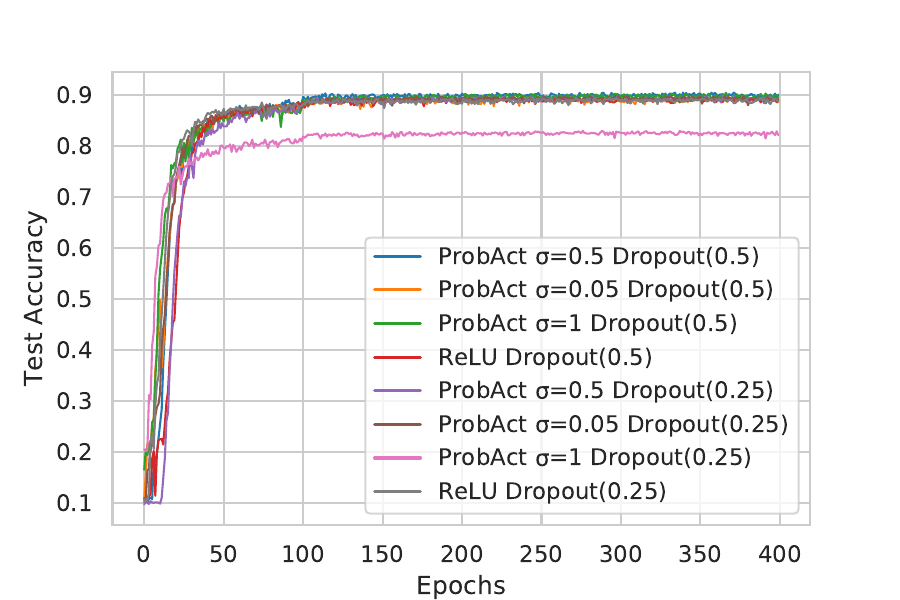}}\hfill
\subfigure[]{\includegraphics[width=0.25\linewidth]{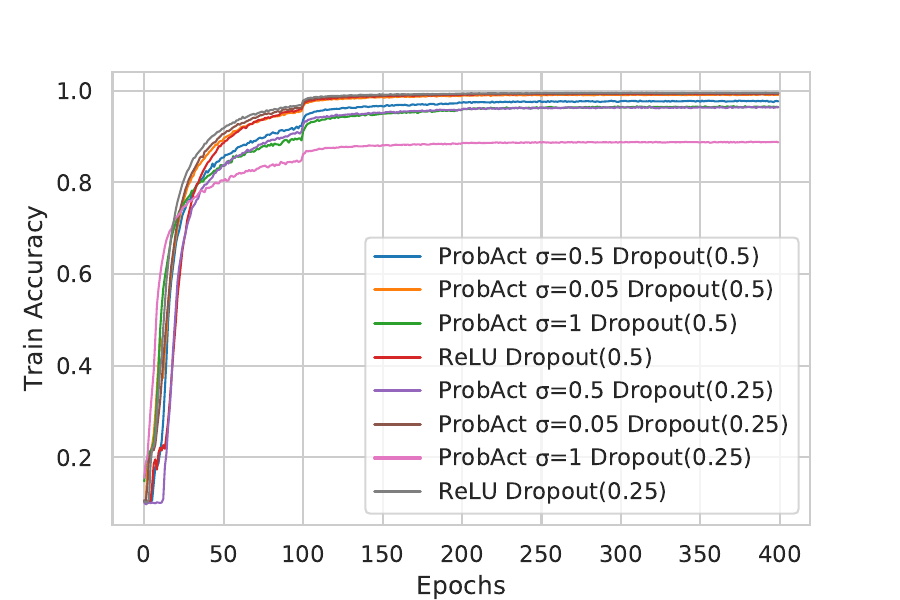}}\hfill
\subfigure[]{\includegraphics[width=0.25\linewidth]{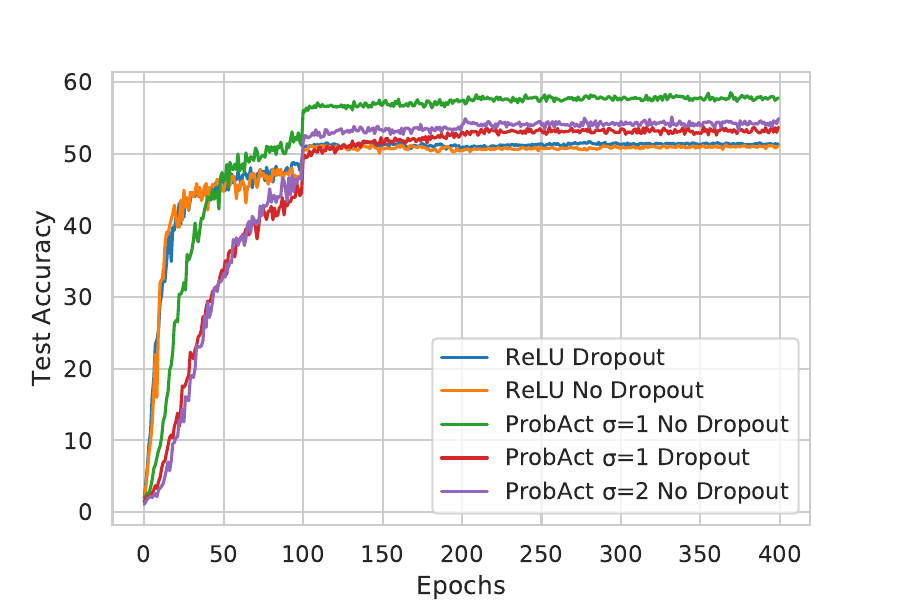}}\hfill
\subfigure[]{\includegraphics[width=0.25\linewidth]{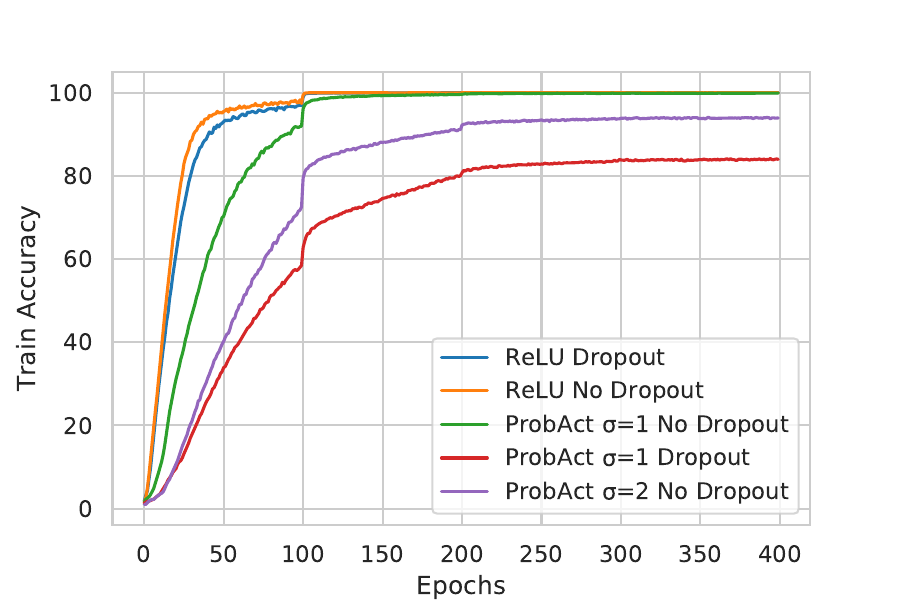}}\hfill
\caption{Training element-wise $\mu$ with a high $\sigma$ value acts an inbuilt regularizer that generalizes better on data, preventing overfitting. We evaluate the network generalization capabilities based on the train and test performance of the network. (a) and (c) shows the test accuracy comparison between ReLU and ProbAct layer with/without dropout layers on CIFAR-10 and CIFAR-100 datasets respectively, while (b) and (d) shows the train accuracy comparison of the same. Network with ProbAct layers achieves better test results and a lower train test accuracy difference, showing the built-in regularization nature of ProbAct.}
\label{fig:comparison}
\end{figure}

\subsection{Uncertainty Estimation}
ProbAct adds stochastic behavior to the neural network and it can be defined as:
\begin{equation}
p(y\ |\ x, \theta) =\displaystyle \int p(y\ |\ x, \theta, \epsilon)\ p(\epsilon)\ d\epsilon =\mathbb{E}_{p(\epsilon)} f(x,\theta,\epsilon)
\end{equation}
where, $y$ : predicted output;  $x$: input;  $\theta$: model parameters;  $\epsilon$: noise; $X$: mini batch data.

\begin{figure}
\subfigure[Predictive uncertainty estimation]{\includegraphics[width=0.33\linewidth]{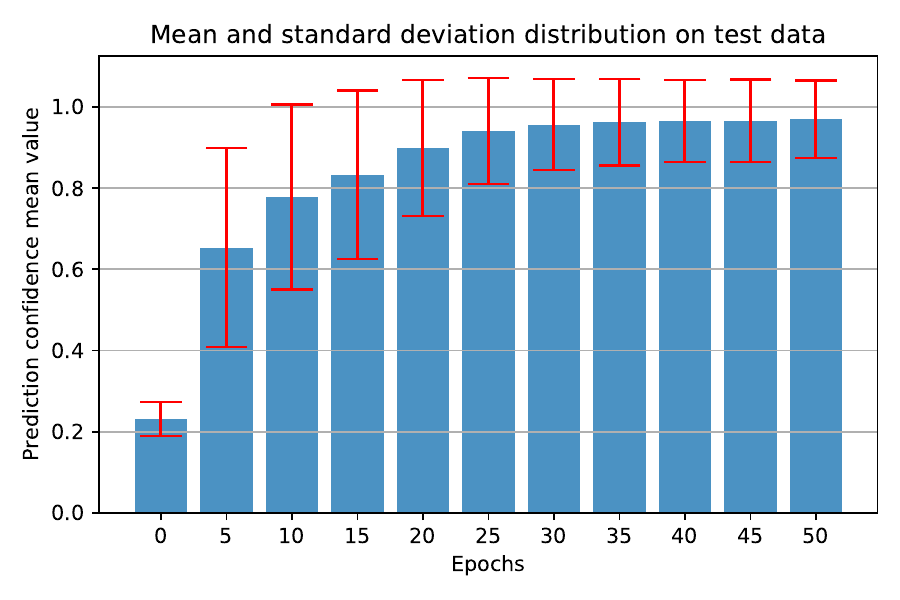}}\hfill
\subfigure[Mean and Standard Deviation]{\includegraphics[width=0.33\linewidth]{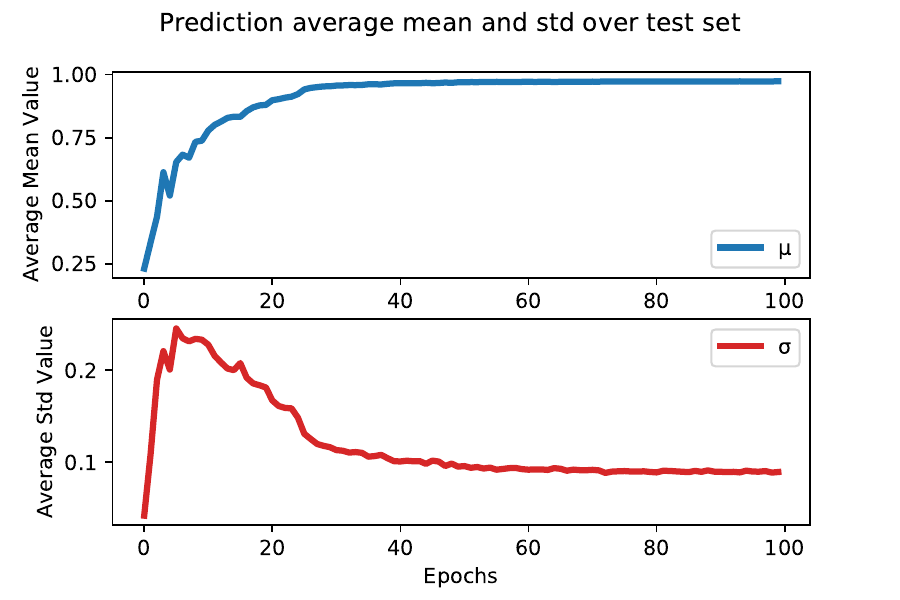}}
\subfigure[Noisy Input prediction]{\includegraphics[width=0.27\linewidth]{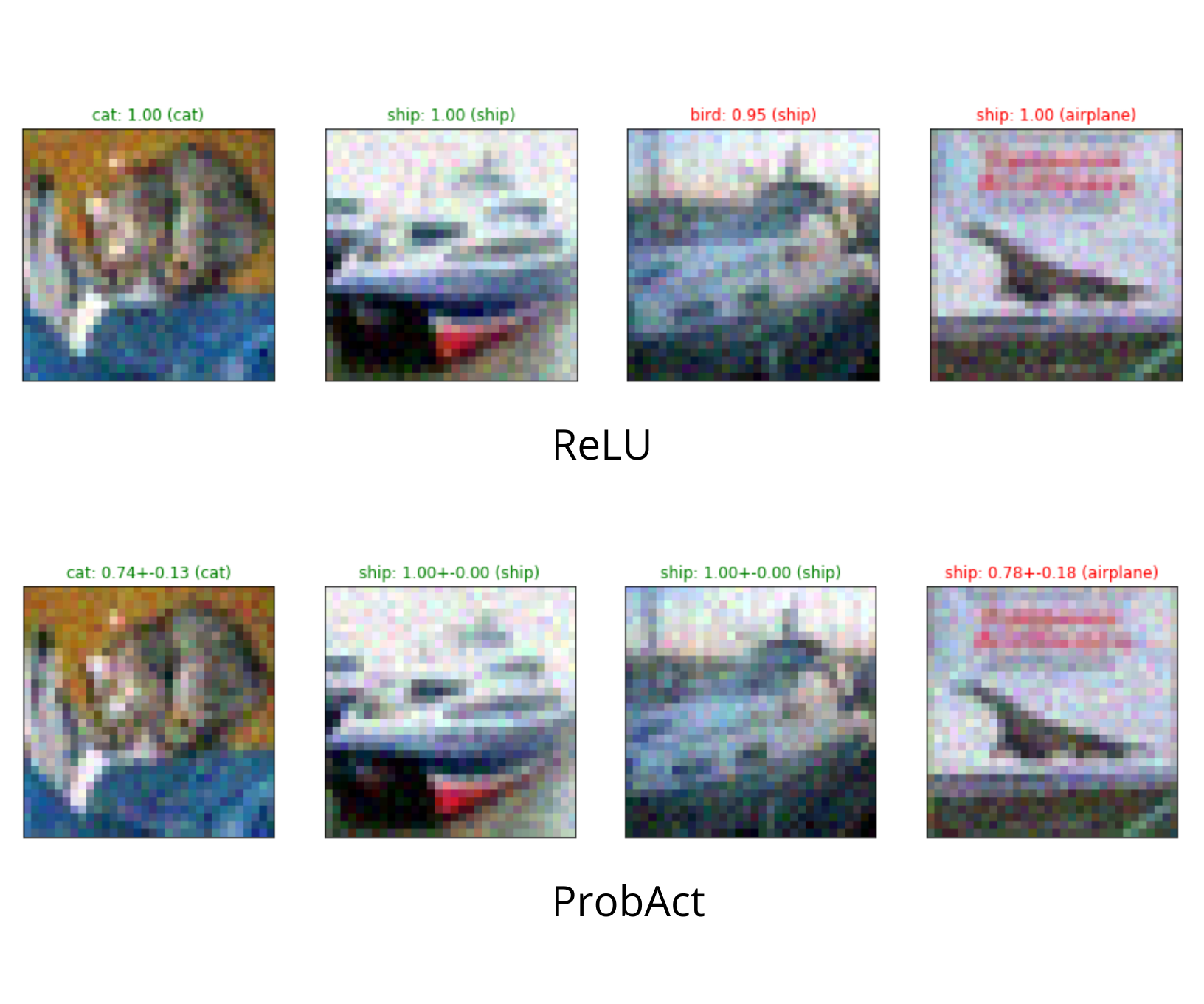}}
\caption{(a) shows the predictive mean (blue bar) and the standard deviation ( error margins in red). With more data, the model uncertainty decreases (clearly seen with increasing mean and decreasing variance). (b) demonstrates the increase in predictive confidence mean is directly proportional to the decrease in the variance. (c) shows that for ProbAct prevents over confident decisions on noisy inputs.}
\label{fig:mean_var}
\end{figure}

ProbAct learns a Gaussian distribution for every element  over time. Each sample from this distribution propagates a new output to the next layers, eventually allowing ProbAct to learn an infinite ensemble of networks. With these ensemble networks, the predictive mean and variance can be estimated by calculating the first and second-order moments. We also observe a decrease in the predictive uncertainty as the network becomes more confident with its decision as shown in Figure~\ref{fig:mean_var} (a) and (b) for CIFAR-10. We also show that this predictive uncertainty helps in preventing overconfident predictive decisions for noisy data. Figure~\ref{fig:mean_var} (c) shows that for noisy inputs, instead of providing a fixed confident predictive score for a misclassification, ProbAct provides a range of low predictive confidence score, demonstrating its non-confident behaviour. 

\subsection{Reduced Data}
The training sample size was reduced to 50\% and 25\% of the original data size for CIFAR-10 and CIFAR-100 dataset. We maintained the class distribution by randomly choosing 25\% and 50\% images for each class. The process was repeated three times to create three randomly chosen datasets. We run our experiments on all three datasets and average the results. 

 Table~\ref{comparison-table} shows the test accuracy for ReLU and ProbAct with Element-wise Trainable $\sigma$ (bound) on 25\% and 50\% data size. We achieve 3\% average increase in test accuracy when the data size was halved and 2.5\% increase when it was further halved. The higher test accuracy of ProbAct shows the applications of ProbAct in real-life use cases when the training data size is small.

\begin{table}[b]
  \caption{Test Accuracy (\%) comparison between ReLU and ProbAct on reduced subsets of CIFAR-10 and CIFAR-100 (50\% and 25\% of original dataset).The test accuracy(\%) indicates the average of three sets of testing.}
  \label{comparison-table}
  \centering
  \scalebox{0.9}{
  \begin{tabular}{lcccc}
    \toprule            
    Activation function & CIFAR-10 (50\%) & CIFAR-100 (50\%) & CIFAR-10 (25\%) & CIFAR-100 (25\%) \\
    \midrule
    ReLU & 82.74 & 42.36 & 75.62 & 30.42 \\
    ProbAct & \textbf{84.73} & \textbf{46.11} & \textbf{79.02} & \textbf{31.67}\\
    \bottomrule
  \end{tabular}
  }
\end{table}

\section{Conclusion}
In this paper, we introduced a novel probabilistic activation function, ProbAct, that adds perturbation in every activation map, allowing better network generalization capabilities. Through experiments we verified that the stochastic perturbation prevents the network from memorizing the training samples, resulting in evenly optimized network weights and a more robust network with a lower generalization error. Furthermore, we confirmed that the augmentation-like operation in ProbAct is effective for classification tasks even when the number of data points is very less. Finally, we demonstrate that ProbAct is very robust to noisy input data and provides an estimate for predictive uncertainty. 

\section*{Acknowledgement}
We would like to acknowledge the members of Uchida Lab, Fukuoka and Mr. Felix Laumann for their invaluable discussions.  

\section{Appendix}
\label{app:vgg}

The VGG-16 architecture used in the experiments is defined as follows:

\paragraph{VGG16}: $[64, 64, M, 128, 128, M, 256, 256, 256, M, 512, 512, 512, M, 512, 512, 512, M, C]$

where, numbers 64,128 and 256 represents the filters of \emph{Convolution layer} which is followed by a {Batch Normalization layer}, followed by an activation function. \emph{M} represents the \emph{Max Pooling layer} and \emph{C} represents the \emph{Linear classification layer} of dimension (512, number of classes).

Other hyper-parameters settings include:
\begin{table}[!h]
  \caption{Hyper-parameters for the experiments}
  \label{hyper-parameters}
  \centering
  \begin{tabular}{ll}
    \toprule                 \\
    Hyper-parameter & Value  \\
    \midrule
    Convolution Kernel Size & 3     \\
    Convolution layer Padding & 1  \\
    Max-Pooling Kernel Size & 2    \\
    Max-Pooling Stride & 2    \\
    Optimizer & Adam \\
    Batch Size & 256 \\
    Fixed $\sigma$ values &  [0.05, 0.1, 0.25, 0.5, 1, 2]\\
    Learning Rate & 0.01 (Dropped 1/10 after every 100 epochs)\\
    Number of Epochs & 400\\
    Image Resolution & $32\times 32$\\
    Single trainable $\sigma$ Initializer & Zero\\
    Element-wise trainable $\sigma$ Initializer & Xavier initialization\\
    \bottomrule
  \end{tabular}
\end{table}

\subsection{Datasets}

\subsubsection{Image Datasets}
\paragraph{CIFAR-10 Dataset}
The CIFAR-10 dataset consists of 60,000 images with 10 classes, with 6,000 images per class, each image 32 by 32 pixels. The dataset is split into 50,000 training images and 10,000 test images. 
\paragraph{CIFAR-100 Dataset} 
CIFAR-100 dataset has 100 classes containing 600 images per class. There are 500 training images and 100 test images per class. The resolution of the images is also 32 by 32 pixels.
\paragraph{STL-10 Dataset} 
STL-10 dataset has 500 images per class with 10 classes and 100 test images per class. The images are 96 by 96 pixels per image. 

\subsubsection{Text Dataset}
\paragraph{Large Movie Review Dataset} 
Large Movie Review \cite{maas-EtAl:2011:ACL-HLT2011} is a binary dataset for sentiment classification (positive or negative) consisting of 25,000 highly polar movie reviews for training, and 25,000 reviews for testing. 

\subsection{Proofs}

\begin{theorem}
The gradient propagation of a stochastic unit $h$ based on a deterministic function $g$ with inputs $\bm{x}$ (a vector containing outputs from other neurons), internal parameters {$\bm{\phi}$} (weights and bias) and noise $z$ is possible, if $g(\bm{x}, \bm{\phi}, z)$ has non-zero gradients with respect to $\bm{x}$ and $\bm{\phi}$. \cite{bengio2013estimating}
\begin{equation}
\label{eq:theorem 1}
  h = g(\bm{x}, \bm{\phi}, z)
\end{equation}
\end{theorem}

Assume a network has two layers with a unit for each layer, the distribution of the second layer's output $y_2$ differs depending on the first and second layer's weights ($w_1$ and $w_2$) and sigmas ($\sigma_1$ and $\sigma_2$) as:
\begin{align}
    y_2 &= \mu \left[ w_2 \mu(w_1 x) + w_2 \sigma_1 \epsilon \right] + \sigma_2 \epsilon \nonumber \\
        &= 
        \begin{cases} 
            \mu \left[ w_2 \mu(w_1 x)) \right] + w_2 \sigma_1 \epsilon + \sigma_2 \epsilon & \text{if $w_2 \mu(w_1 x) + w_2 \sigma_1 \epsilon > 0$}, \\
            \sigma_2 \epsilon & \mathrm{otherwise,}
        \end{cases} \nonumber \\
        & \sim
        \begin{cases} 
            N(\mu \left[ w_2 \mu(w_1 x)) \right], (w_2 \sigma_1)^2 + \sigma_2^2)\\
            N(0, \sigma_2^2).
        \end{cases}
\end{align}
Incidentally, as shown in Figure 1 (c) in the main paper, a small noise variance tends to be learned in the final layer to make the network output stable (see Section 4.3 in the paper for a quantitative evaluation).

Using  Eq.~\eqref{eq:theorem 1} from Theorem 1,  assume $g$ is noise injection function that depends on noise $z$, and some differentiable transformations $\bm{d}$ over inputs $\bm{x}$ and model internal parameters $\bm{\phi}$.
We can derive the output, $h$ as:
\begin{equation}
\label{eq:stochastic h}
    h= g(\bm{d}, \bm{z})
\end{equation}

If we use Eq.~\eqref{eq:stochastic h} for another noise addition methods like dropout~\cite{hinton2012improving} or masking the noise in denoising auto-encoders \cite{vincent2008extracting}, we can infer $z$ as noise multiplied just after a non-linearity is induced in a neuron. In the case of ProbAct, we sample from Gaussian noise and add it while computing $h$. Or we can say, we add a noise to the pre-activation, which is used as an input to the next layer. In doing so, self regularization behaviour is induced in the network.

\begin{figure}
\subfigure[]{\includegraphics[width=0.5\linewidth]{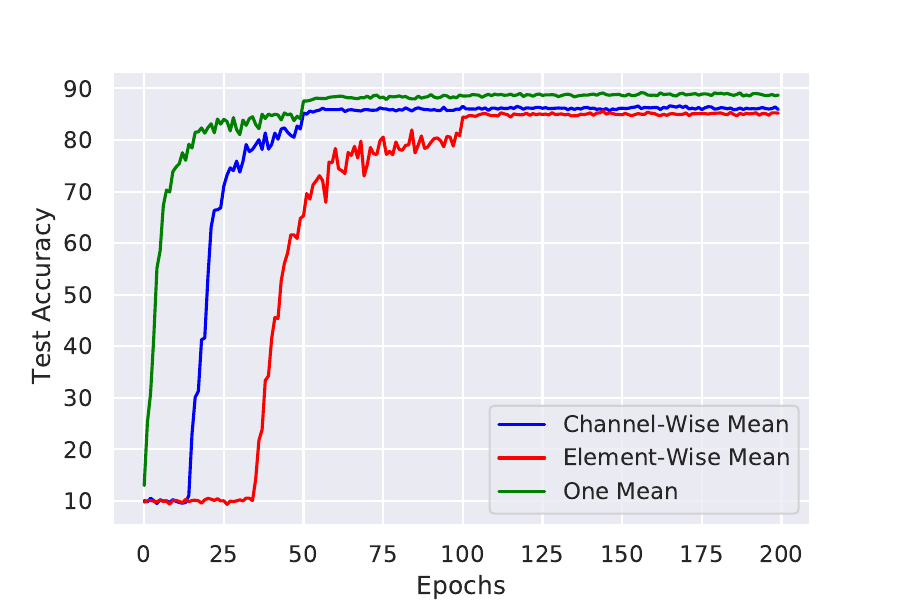}}\hfill
\subfigure[]{\includegraphics[width=0.5\linewidth]{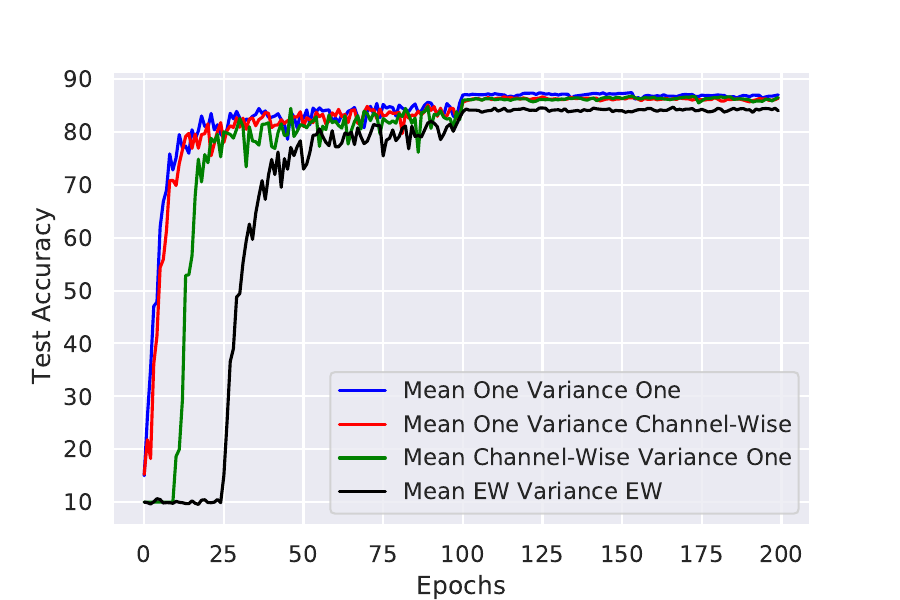}}\hfill
\caption{(a) Comparison of different mean settings with fixed $\sigma=1$ for CIFAR-10 datasets. Channel-Wise mean denotes a constant learnable mean within the channels, element-wise mean denotes learnable mean for each parameter, while one mean denotes a single learnable mean for entire dataset. It is worth noting that element-wise mean takes a lot more time to converge compared to one mean or channel-wise mean. This shows that using single learnable mean uses fewer parameters and converges faster.(b) denotes the different variance settings for CIFAR-10 dataset. On contrary to mean settings, there is no clear convergence time difference between single variance and channel-wise variance. Single variance is preferred due to the usage of fewer parameters.Training element-wise mean and element-wise variance (denoted as EW) leads to too many learnable parameters and takes a long time to converge. Training time is high for such a solution and is only preferred when overfitting is an issue.}
\label{fig:mean_sigma_comparison}
\end{figure}

\begin{figure}
{\includegraphics[width=1.0\linewidth]{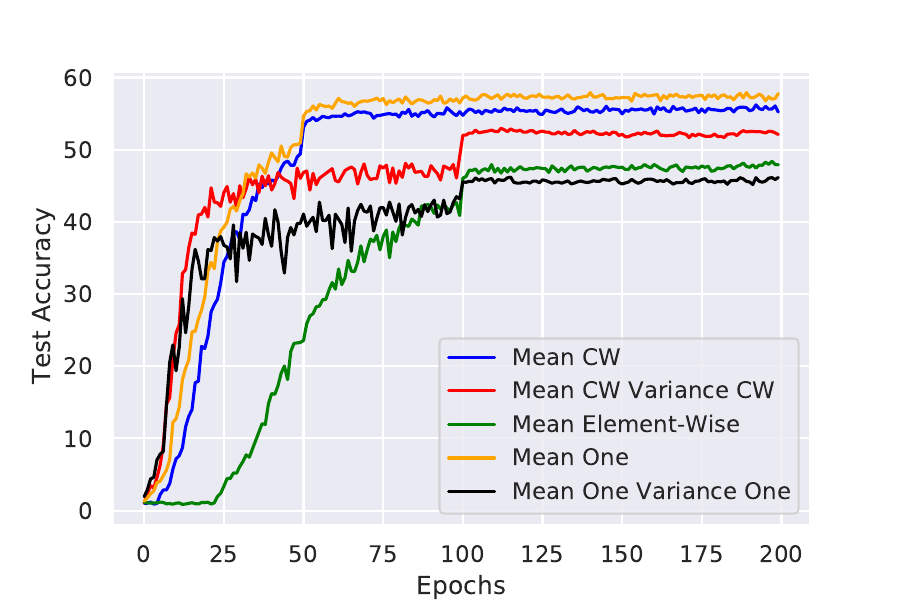}}\hfill
\caption{Different mean and variance setting of ProbAct for CIFAR-100 dataset. It is worth noting that single learnable mean is equally effective compared to channel-wise learnable mean but with fewer parameters. Element-wise mean accuracy is lesser than the others but it is more susceptible to overfitting as stated in the paper.}
\label{fig:cifar100_comparison}
\end{figure}

\begin{figure}
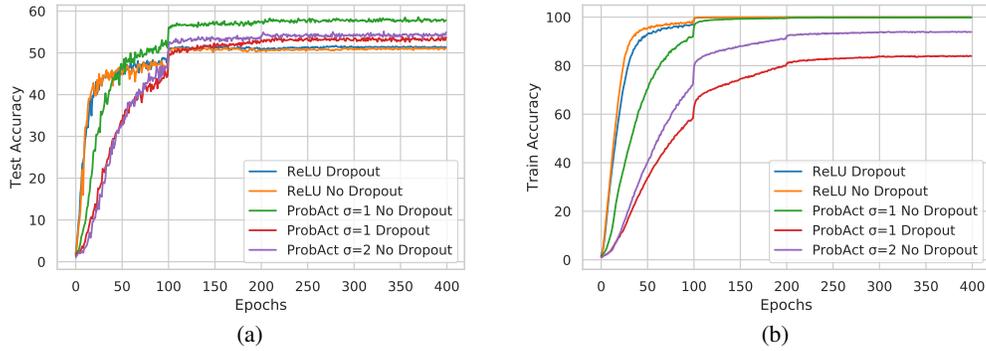

\subfigure[]{\includegraphics[width=0.5\linewidth]{TestAcc.pdf}}\hfill
\subfigure[]{\includegraphics[width=0.5\linewidth]{TrainAcc.pdf}}\hfill
    \caption{(a) and (b) shows the train and test accuracy comparison between ReLU and ProbAct layer with/without dropout layers on CIFAR-100 dataset. With $\sigma = 2$, ProbAct achieves similar test performance to ReLU activation layer with dropout with less overfitting as shown from the training curve. Adding a dropout layer further improves the generalization capabilities, showing the built-in regularization nature of ProbAct.}
\label{fig:OverfittingCIFAR100comparison}
\end{figure}

\begin{figure}[t] 
\centering
\includegraphics[width=1\linewidth]{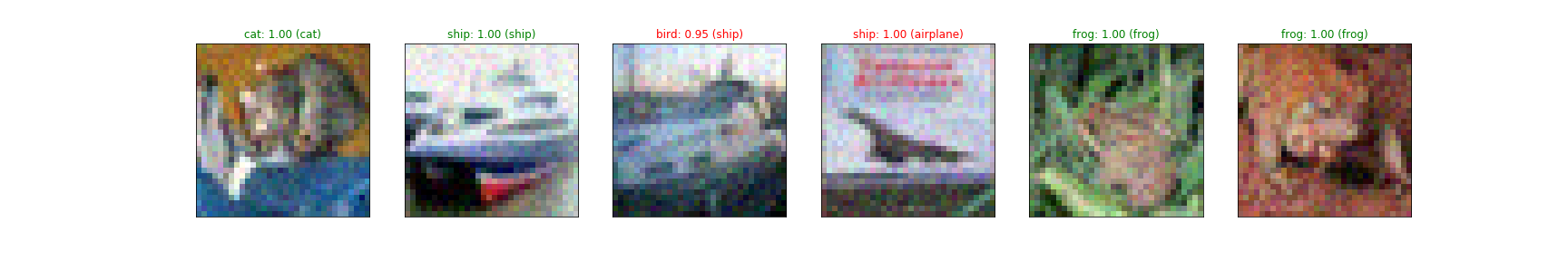}
\includegraphics[width=1\linewidth]{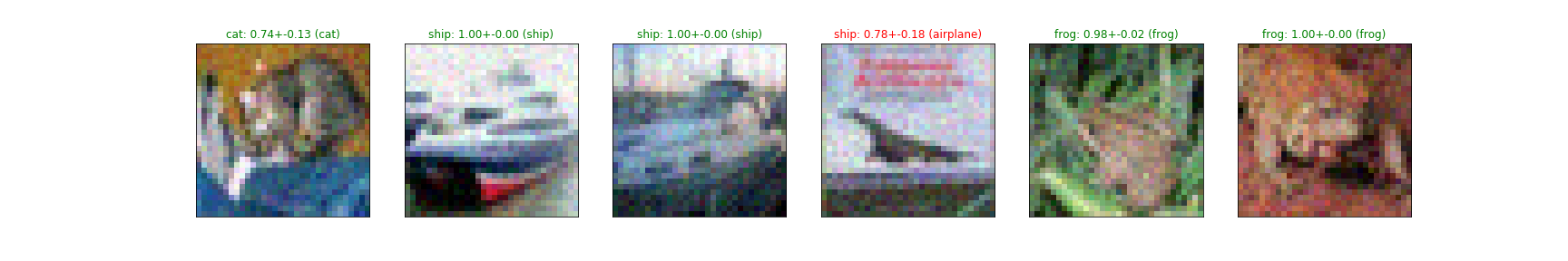}
\caption{Comparison of predictive test accuracy and confidence score between ReLU (above) and ProbAct (below) on a VGG-16 architecture on CIFAR-10 dataset with 0.05 Gaussian noise added to test samples.}
\label{fig:cifar-10testnoise}
\end{figure}



\bibliographystyle{IEEEtran}
\bibliography{ref}

\begin{thebibliography}{10}
\providecommand{\url}[1]{#1}
\csname url@samestyle\endcsname
\providecommand{\newblock}{\relax}
\providecommand{\bibinfo}[2]{#2}
\providecommand{\BIBentrySTDinterwordspacing}{\spaceskip=0pt\relax}
\providecommand{\BIBentryALTinterwordstretchfactor}{4}
\providecommand{\BIBentryALTinterwordspacing}{\spaceskip=\fontdimen2\font plus
\BIBentryALTinterwordstretchfactor\fontdimen3\font minus
  \fontdimen4\font\relax}
\providecommand{\BIBforeignlanguage}[2]{{%
\expandafter\ifx\csname l@#1\endcsname\relax
\typeout{** WARNING: IEEEtran.bst: No hyphenation pattern has been}%
\typeout{** loaded for the language `#1'. Using the pattern for}%
\typeout{** the default language instead.}%
\else
\language=\csname l@#1\endcsname
\fi
#2}}
\providecommand{\BIBdecl}{\relax}
\BIBdecl

\bibitem{article2}
A.~Vehbi~Olgac and B.~Karlik, ``Performance analysis of various activation
  functions in generalized {MLP} architectures of neural networks,''
  \emph{International Journal of Artificial Intelligence And Expert Systems},
  vol.~1, pp. 111--122, 02 2011.

\bibitem{Cybenko1989}
G.~Cybenko, ``Approximation by superpositions of a sigmoidal function,''
  \emph{Mathematics of Control, Signals, and Systems}, vol.~2, no.~4, pp.
  303--314, dec 1989.

\bibitem{schmidhuber2015deep}
J.~Schmidhuber, ``Deep learning in neural networks: An overview,'' \emph{Neural
  Networks}, vol.~61, pp. 85--117, 2015.

\bibitem{nair2010rectified}
V.~Nair and G.~E. Hinton, ``Rectified linear units improve restricted boltzmann
  machines,'' in \emph{International Conference on Machine Learning}, 2010, pp.
  807--814.

\bibitem{xu2015empirical}
B.~Xu, N.~Wang, T.~Chen, and M.~Li, ``Empirical evaluation of rectified
  activations in convolutional network,'' \emph{arXiv preprint
  arXiv:1505.00853}, 2015.

\bibitem{he2015delving}
K.~He, X.~Zhang, S.~Ren, and J.~Sun, ``Delving deep into rectifiers: Surpassing
  human-level performance on {ImageNet} classification,'' in
  \emph{International Conference on Computer Vision}, dec 2015.

\bibitem{2015arXiv151107289C}
D.-A. Clevert, T.~Unterthiner, and S.~Hochreiter, ``Fast and accurate deep
  network learning by exponential linear units ({ELUs}),'' \emph{arXiv preprint
  arXiv:1511.07289}, 2015.

\bibitem{lewicki1998review}
M.~S. Lewicki, ``A review of methods for spike sorting: the detection and
  classification of neural action potentials,'' \emph{Network: Computation in
  Neural Systems}, vol.~9, no.~4, pp. R53--R78, 1998.

\bibitem{Ramachandran2018SearchingFA}
P.~Ramachandran, B.~Zoph, and Q.~V. Le, ``Searching for activation functions,''
  \emph{arXiv preprint arXiv:1710.05941}, 2017.

\bibitem{Basirat2018TheQF}
M.~Basirat and P.~M. Roth, ``The quest for the golden activation function,''
  \emph{arXiv preprint arXiv:1808.00783}, 2018.

\bibitem{Jin:2016:DLS:3016100.3016142}
X.~Jin, C.~Xu, J.~Feng, Y.~Wei, J.~Xiong, and S.~Yan, ``Deep learning with
  s-shaped rectified linear activation units,'' in \emph{AAAI Conference on
  Artificial Intelligence}, 2016, pp. 1737--1743.

\bibitem{trottier2017parametric}
L.~Trottier, P.~Gigu, B.~Chaib-draa \emph{et~al.}, ``Parametric exponential
  linear unit for deep convolutional neural networks,'' in \emph{IEEE
  International Conference on Machine Learning and Applications}, 2017, pp.
  207--214.

\bibitem{pmlr-v48-gulcehre16}
C.~Gulcehre, M.~Moczulski, M.~Denil, and Y.~Bengio, ``Noisy activation
  functions,'' in \emph{International Conference on Machine Learning}, vol.~48,
  jun 2016, pp. 3059--3068.

\bibitem{srivastava2014dropout}
N.~Srivastava, G.~Hinton, A.~Krizhevsky, I.~Sutskever, and R.~Salakhutdinov,
  ``Dropout: a simple way to prevent neural networks from overfitting,''
  \emph{Journal of Machine Learning Research}, vol.~15, no.~1, pp. 1929--1958,
  2014.

\bibitem{liu2018towards}
X.~Liu, M.~Cheng, H.~Zhang, and C.-J. Hsieh, ``Towards robust neural networks
  via random self-ensemble,'' in \emph{European Conference on Computer Vision},
  2018, pp. 369--385.

\bibitem{Inayoshi}
H.~Inayoshi and T.~Kurita, ``Improved generalization by adding both
  auto-association and hidden-layer-noise to
  neural-network-based-classifiers,'' in \emph{{IEEE} Workshop on Machine
  Learning for Signal Processing}, 2005.

\bibitem{Bishop_1995}
C.~M. Bishop, ``Training with noise is equivalent to tikhonov regularization,''
  \emph{Neural Computation}, vol.~7, no.~1, pp. 108--116, jan 1995.

\bibitem{An_1996}
G.~An, ``The effects of adding noise during backpropagation training on a
  generalization performance,'' \emph{Neural Computation}, vol.~8, no.~3, pp.
  643--674, apr 1996.

\bibitem{Audhkhasi_2013}
K.~Audhkhasi, O.~Osoba, and B.~Kosko, ``Noise benefits in backpropagation and
  deep bidirectional pre-training,'' in \emph{International Joint Conference on
  Neural Networks}, aug 2013.

\bibitem{neelakantan2015adding}
A.~Neelakantan, L.~Vilnis, Q.~V. Le, I.~Sutskever, L.~Kaiser, K.~Kurach, and
  J.~Martens, ``Adding gradient noise improves learning for very deep
  networks,'' \emph{arXiv preprint arXiv:1511.06807}, 2015.

\bibitem{NIPS2011_4329}
A.~Graves, ``Practical variational inference for neural networks,'' in
  \emph{Advances in Neural Information Processing Systems}, J.~Shawe-Taylor,
  R.~S. Zemel, P.~L. Bartlett, F.~Pereira, and K.~Q. Weinberger, Eds., 2011,
  pp. 2348--2356.

\bibitem{Murray_1993}
A.~Murray and P.~Edwards, ``Synaptic weight noise during multilayer perceptron
  training: fault tolerance and training improvements,'' \emph{{IEEE}
  Transactions on Neural Networks}, vol.~4, no.~4, pp. 722--725, jul 1993.

\bibitem{pmlr-v37-blundell15}
C.~Blundell, J.~Cornebise, K.~Kavukcuoglu, and D.~Wierstra, ``Weight
  uncertainty in neural network,'' in \emph{International Conference on Machine
  Learning}, vol.~37, 2015, pp. 1613--1622.

\bibitem{bengio2013estimating}
Y.~Bengio, N.~L{\'e}onard, and A.~Courville, ``Estimating or propagating
  gradients through stochastic neurons for conditional computation,''
  \emph{arXiv preprint arXiv:1308.3432}, 2013.

\bibitem{mackay1992practical}
D.~J. MacKay, ``A practical bayesian framework for backpropagation networks,''
  \emph{Neural computation}, vol.~4, no.~3, pp. 448--472, 1992.

\bibitem{graves2011practical}
A.~Graves, ``Practical variational inference for neural networks,'' in
  \emph{Advances in neural information processing systems}, 2011, pp.
  2348--2356.

\bibitem{hinton1993keeping}
G.~Hinton and D.~Van~Camp, ``Keeping neural networks simple by minimizing the
  description length of the weights,'' in \emph{in Proc. of the 6th Ann. ACM
  Conf. on Computational Learning Theory}.\hskip 1em plus 0.5em minus
  0.4em\relax Citeseer, 1993.

\bibitem{shridhar2019comprehensive}
K.~Shridhar, F.~Laumann, and M.~Liwicki, ``A comprehensive guide to bayesian
  convolutional neural network with variational inference,'' \emph{arXiv
  preprint arXiv:1901.02731}, 2019.

\bibitem{kendall2017uncertainties}
A.~Kendall and Y.~Gal, ``What uncertainties do we need in bayesian deep
  learning for computer vision?'' in \emph{Advances in neural information
  processing systems}, 2017, pp. 5574--5584.

\bibitem{kwon2018uncertainty}
Y.~Kwon, J.-H. Won, B.~Kim, and M.~Paik, ``Uncertainty quantification using
  bayesian neural networks in classification: Application to ischemic stroke
  lesion segmentation,'' \emph{Computational Statistics and Data Analysis}, 04
  2018.

\bibitem{shridhar2018uncertainty}
K.~Shridhar, F.~Laumann, and M.~Liwicki, ``Uncertainty estimations by softplus
  normalization in bayesian convolutional neural networks with variational
  inference,'' \emph{arXiv preprint arXiv:1806.05978}, 2018.

\bibitem{lakshminarayanan2017simple}
B.~Lakshminarayanan, A.~Pritzel, and C.~Blundell, ``Simple and scalable
  predictive uncertainty estimation using deep ensembles,'' in \emph{Advances
  in neural information processing systems}, 2017, pp. 6402--6413.

\bibitem{DBLP:journals/corr/WangSY16}
\BIBentryALTinterwordspacing
H.~Wang, X.~Shi, and D.~Yeung, ``Natural-parameter networks: {A} class of
  probabilistic neural networks,'' \emph{CoRR}, vol. abs/1611.00448, 2016.
  [Online]. Available: \url{http://arxiv.org/abs/1611.00448}
\BIBentrySTDinterwordspacing

\bibitem{DBLP:journals/corr/abs-1805-11327}
\BIBentryALTinterwordspacing
J.~Gast and S.~Roth, ``Lightweight probabilistic deep networks,'' \emph{CoRR},
  vol. abs/1805.11327, 2018. [Online]. Available:
  \url{http://arxiv.org/abs/1805.11327}
\BIBentrySTDinterwordspacing

\bibitem{hernandez2015probabilistic}
J.~M. Hern{\'a}ndez-Lobato and R.~Adams, ``Probabilistic backpropagation for
  scalable learning of bayesian neural networks,'' in \emph{International
  Conference on Machine Learning}, 2015, pp. 1861--1869.

\bibitem{cifar10}
\BIBentryALTinterwordspacing
A.~Krizhevsky, V.~Nair, and G.~Hinton, ``Cifar-10 (canadian institute for
  advanced research).'' [Online]. Available:
  \url{http://www.cs.toronto.edu/~kriz/cifar.html}
\BIBentrySTDinterwordspacing

\bibitem{coates2011analysis}
A.~Coates, A.~Ng, and H.~Lee, ``An analysis of single-layer networks in
  unsupervised feature learning,'' in \emph{International Conference on
  Artificial Intelligence and Statistics}, 2011, pp. 215--223.

\bibitem{maas-EtAl:2011:ACL-HLT2011}
\BIBentryALTinterwordspacing
A.~L. Maas, R.~E. Daly, P.~T. Pham, D.~Huang, A.~Y. Ng, and C.~Potts,
  ``Learning word vectors for sentiment analysis,'' in \emph{Proceedings of the
  49th Annual Meeting of the Association for Computational Linguistics: Human
  Language Technologies}.\hskip 1em plus 0.5em minus 0.4em\relax Portland,
  Oregon, USA: Association for Computational Linguistics, June 2011, pp.
  142--150. [Online]. Available: \url{http://www.aclweb.org/anthology/P11-1015}
\BIBentrySTDinterwordspacing

\bibitem{clevert2015fast}
D.-A. Clevert, T.~Unterthiner, and S.~Hochreiter, ``Fast and accurate deep
  network learning by exponential linear units (elus),'' \emph{arXiv preprint
  arXiv:1511.07289}, 2015.

\bibitem{klambauer1706self}
G.~Klambauer, T.~Unterthiner, A.~Mayr, and S.~Hochreiter, ``Self-normalizing
  neural networks. arxiv 2017,'' \emph{arXiv preprint arXiv:1706.02515}, 2017.

\bibitem{Shridhar2018BayesianCN}
\BIBentryALTinterwordspacing
F.~Laumann, K.~Shridhar, and A.~L. Maurin, ``Bayesian convolutional neural
  networks,'' \emph{CoRR}, vol. abs/1806.05978, 2018. [Online]. Available:
  \url{http://arxiv.org/abs/1806.05978}
\BIBentrySTDinterwordspacing

\bibitem{2014arXiv1409.1556S}
K.~Simonyan and A.~Zisserman, ``Very deep convolutional networks for
  large-scale image recognition,'' \emph{arXiv preprint arXiv:1409.1556}, 2014.

\bibitem{hinton2012improving}
G.~E. Hinton, N.~Srivastava, A.~Krizhevsky, I.~Sutskever, and R.~R.
  Salakhutdinov, ``Improving neural networks by preventing co-adaptation of
  feature detectors,'' \emph{arXiv preprint arXiv:1207.0580}, 2012.

\bibitem{vincent2008extracting}
P.~Vincent, H.~Larochelle, Y.~Bengio, and P.-A. Manzagol, ``Extracting and
  composing robust features with denoising autoencoders,'' in \emph{ACM
  International Conference on Machine Learning}, 2008, pp. 1096--1103.

\end{thebibliography}

\end{document}